\documentclass[sigconf]{acmart}
\AtBeginDocument{%
  }

\usepackage{booktabs}
\usepackage{multirow}
\usepackage{colortbl}
\usepackage{xcolor}
\usepackage{pifont}
\usepackage{tikz}
\usepackage[misc]{ifsym}
\usepackage{caption}
\usepackage{subcaption}
\usepackage{enumitem}
\usepackage{algorithm}
\usepackage{algorithmic}

\definecolor{mygray}{gray}{.85}
\definecolor{myred}{RGB}{220,36,36}
\definecolor{myblue}{RGB}{0,51,160}

\newcommand{\best}[1]{\cellcolor{red!15}\textbf{#1}}
\newcommand{\second}[1]{\cellcolor{green!15}#1}

\newcommand{\model}{EMG}

\copyrightyear{2027}
\acmYear{2027}
\acmPrice{}
\acmDOI{10.1145/......}
\acmISBN{......}

\begin{document}


\title{Experience Memory Graph: Extracting Reusable Correction Knowledge from Agent Trajectories}

\author{Wenjun Wang}
\authornote{Equal Contribution.}
\affiliation{%
  \institution{University of Electronic Science and Technology of China}
  \city{Chengdu}
  \country{China}
}
\email{202522080926@std.uestc.edu.cn}

\author{Yuchen Fang}
\authornotemark[1]
\affiliation{%
  \institution{University of Electronic Science and Technology of China}
  \city{Chengdu}
  \country{China}
}
\email{fangyuchen@std.uestc.edu.cn}

\author{Fengrui Liu}
\affiliation{%
  \institution{University of Electronic Science and Technology of China}
  \city{Chengdu}
  \country{China}
}
\email{202522080925@std.uestc.edu.cn}

\author{Zibo Liang}
\affiliation{%
  \institution{University of Electronic Science and Technology of China}
  \city{Chengdu}
  \country{China}
}
\email{zbliang@std.uestc.edu.cn}


\author{Kai Zheng}
\authornote{Corresponding author: Kai Zheng. Kai Zheng is with Yangtze Delta Region Institute (Quzhou), School of Computer Science and Engineering, UESTC. He is also with Shenzhen Institute for Advanced Study, UESTC.}
\affiliation{%
  \institution{University of Electronic Science and Technology of China}
  \city{Chengdu}
  \country{China}
}
\email{zhengkai@uestc.edu.cn}

\renewcommand{\shortauthors}{Wenjun Wang et al.}

\begin{abstract}
Large Language Model (LLM) agents have shown remarkable capabilities in autonomous decision-making by generating sequential trajectories of states, actions, and observations. However, in complex, long-horizon tasks, these agents frequently suffer from compounding errors and struggle to recover from failures. Despite advances in experience memory, extracting reliable and transferable correction knowledge from historical trajectories remains challenging. Existing methods, whether based on test-time reflection or offline memory construction, largely rely on LLMs to diagnose failures and formulate corrective strategies, which can produce brittle or task-specific insights that are difficult to generalize. To address this challenge, we propose Experience Memory Graph (EMG), a framework that reformulates agent failure recovery as a graph matching problem. During offline construction on the training set, we convert both failed exploration trajectories and successful expert trajectories into directed action decision graphs. By matching these graphs, we extract common subgraphs and graph edit paths that explicitly indicate how to correct failures, and store them in a memory graph with intra-task nodes and cross-task edges. At test time, EMG retrieves relevant insights and guides the agent in a single test-time attempt without additional reflection-and-replay rounds. Experiments on ALFWorld and ScienceWorld demonstrate consistent improvements over the evaluated memory construction and competitive baselines across seen and unseen splits.
\end{abstract}



\maketitle

\section{Introduction}

Large Language Model (LLM) agents are increasingly being developed to interact with Web pages and online services, enabling complex tasks that require multi-step planning, sequential actions, and decision-making based on environmental feedback~\cite{1,2,3}. Such tasks involve interdependent decisions, where errors in earlier steps can propagate and prevent task completion. Experience memory offers a promising way to improve agent reliability by retaining useful knowledge from historical interactions and reusing it in future tasks~\cite{12,5,34}. However, effectively leveraging both successful and failed trajectories requires distinguishing reusable behaviors from erroneous decisions and translating these experiences into actionable guidance for subsequent executions.

Existing methods exploit historical experience through failure diagnosis, reflective learning, and memory construction~\cite{6,7,8,11,13,16,20,22,24,37,39}. Demonstration- and workflow-based approaches reuse successful trajectories or extract reusable action routines, as in Synapse~\cite{18} and Agent Workflow Memory~\cite{5}. Reflection-based methods use execution feedback to improve subsequent behavior, including Reflexion~\cite{33} and ExpeL~\cite{34}, while related work explores reflective decision-making and experience-driven adaptation~\cite{10,21}. Memory construction methods abstract reusable knowledge from historical interactions: ReasoningBank~\cite{9} extracts reasoning strategies, MemP~\cite{12} distills procedural knowledge, and other approaches explore strategy memory and context-dependent retrieval~\cite{15,24}. A-Mem~\cite{35} links memory notes to support retrieval and reuse, while graph-structured skill composition offers another way to organize reusable capabilities~\cite{14}. Across these directions, LLM prompting is widely used to generate and refine memory content, including workflow summaries, failure reflections, reasoning strategies, and contextualized memory notes.

These prompt-based extraction mechanisms, however, typically leave decision correspondences implicit in the LLM's reasoning. As illustrated in Figure~1(a), prompts guide the model to summarize successful behavior and derive corrective advice from failures, while the relationships between the underlying decisions are inferred during memory generation. Even when successful and failed executions are presented together, the model must determine which observation-conditioned decisions correspond, which behaviors should be preserved, and which differences suggest a need for correction. Similarly, generating and linking related memories does not by itself identify the workflow structures shared by their underlying executions. These limitations motivate explicit structural comparisons within and across tasks, giving rise to two challenges for extracting reusable correction knowledge.

\textbf{Challenge I: Implicit Decision Correspondence.}
A failed trajectory may contain useful intermediate decisions, while a successful trajectory may include incidental or redundant actions. Comparing their outcomes therefore does not directly identify which decisions should be preserved or changed. Long trajectories further contain repeated actions, invalid attempts, and differences in execution order that complicate correspondence. An action must also be interpreted in the context of its preceding observation: the same action may be appropriate in one context and ineffective in another. When these relationships remain implicit, the LLM must infer both the correspondence between decisions and the corrective advice that follows, making the specificity of the extracted knowledge dependent on its ability to compare and diagnose long executions.

\textbf{Challenge II: Limited Structural Evidence for Cross-Task Reuse.}
Existing strategy abstraction and memory linking already facilitate experience reuse across tasks. However, a relevant strategy or memory connection does not necessarily specify which observation--action patterns are shared by the underlying successful executions. Corrective advice extracted from one task can also combine reusable behavior with details specific to its objects, locations, or goal, leaving the agent to determine which parts apply to a new task. Extracting shared workflows from related successful trajectories can provide additional structural evidence for this adaptation, connecting task-level correction knowledge with behavior supported across multiple tasks.

\begin{figure}[t]
    \centering
        \includegraphics[width=\linewidth]{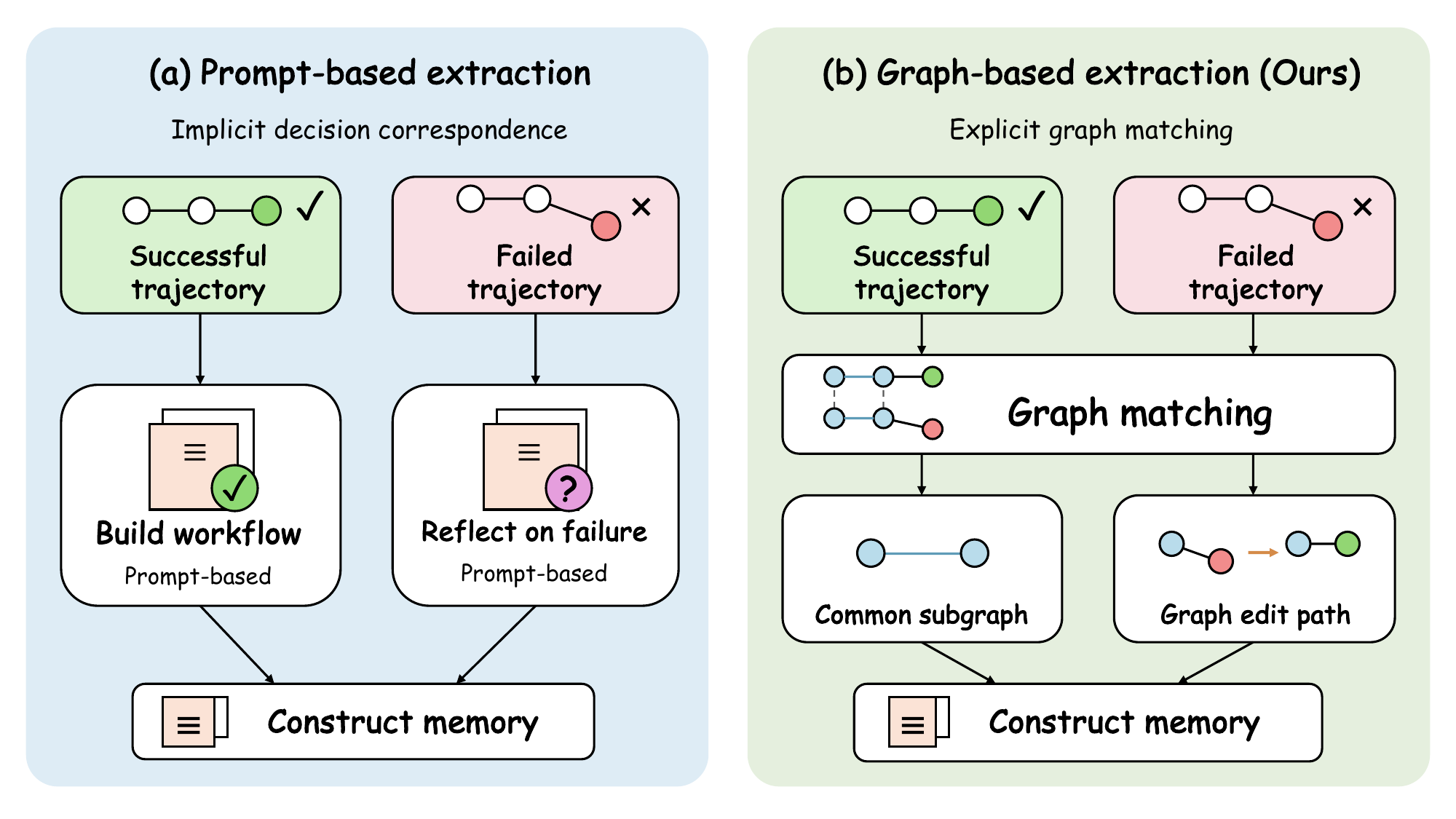}
    \caption{Comparison of experience extraction mechanisms. (a) Prompt-based extraction summarizes successful workflows and failure reflections while leaving decision correspondences implicit. (b) EMG explicitly matches action decision graphs to extract common subgraphs and graph edit paths, providing structural evidence for distilling reusable workflow and correction insights into memory.}
    \label{intro:sketch}
\end{figure}

To address these challenges, we propose \textbf{Experience Memory Graph (EMG)}, a framework that extracts reusable correction knowledge through explicit graph matching, as illustrated in Figure~1(b). For each training task, EMG converts an exploration trajectory and its corresponding successful expert trajectory into directed action decision graphs. The expert trajectories are obtained from the supervised fine-tuning dataset provided by ETO~\cite{17}. Actions form the nodes, while preceding observations label the edges, making the observed conditions associated with action selection explicit. Matching the graphs establishes structural correspondences between the executions. \textit{common subgraphs} expose shared workflow fragments, while \textit{graph edit paths} express differences as operations that transform the failed exploration graph into its successful reference. Depending on the exploration outcome, these structures guide an LLM in distilling workflows, corrective insights, or both. By establishing explicit decision correspondences and providing structural evidence about what can be preserved and what may need to change, this extraction mechanism addresses \textbf{Challenge~I}.

EMG further organizes this knowledge within and across tasks. Insights derived from exploration--expert comparisons are stored on task nodes in a memory graph. For related tasks, EMG matches their expert graphs to derive common subgraphs and extract shared workflows, which are summarized into cross-task insights and stored on memory edges. These shared workflows provide structural evidence for cross-task experience reuse, addressing \textbf{Challenge~II}. Retrieving node and edge insights together provides task-level correction knowledge and complementary guidance from patterns shared across successful executions. The memory is constructed offline using training tasks. At test time, the retrieved insights guide the agent's multi-step interaction within a single task-level attempt, without additional reflection-and-replay rounds.

Our contributions are threefold:
\begin{itemize}
    \item We introduce EMG, a framework that extracts reusable correction knowledge through explicit graph matching. By aligning exploration and expert action decision graphs, EMG derives common subgraphs and graph edit paths that provide structural evidence for generating workflow and correction insights, addressing \textbf{Challenge~I}.

    \item We derive cross-task insights by matching expert graphs from related tasks and extracting their shared workflows. These insights are stored on memory edges and retrieved together with task-level node insights, providing structural context for reusing correction knowledge across tasks and addressing \textbf{Challenge~II}.

    \item Experiments on ALFWorld and ScienceWorld demonstrate consistent improvements over the evaluated memory construction and iterative reflection baselines across seen and unseen splits with two execution models. Ablation studies support the contributions of graph-based extraction and cross-task insights.
\end{itemize}

\section{Related Work}
\subsection{Experience Memory}

Experience memory extracts reusable workflows and insights from historical trajectories. Early methods, such as Agent Workflow Memory \cite{5} and Synapse \cite{18}, primarily derive reusable workflows from successful executions, overlooking the valuable signals contained in failures.

To exploit failure information, some studies focus on diagnosing errors. AgenTracer \cite{6} uses counterfactual replay and fault injection to identify failure causes, while Trajectory Graph Copilot \cite{7} models trajectories as graphs to detect error-prone action patterns. AgentDebug \cite{8} further systematically categorizes agent failures and analyzes their common causes.

Another line of work extracts reusable experience through reflection and memory construction. Test-time methods update memory using feedback from ongoing interactions. R2D2 \cite{10} stores reflections and corrected experiences, SAGE \cite{21} iteratively refines strategies and memory, and CDMem \cite{15} improves context-aware memory encoding and retrieval. Offline methods instead distill reusable knowledge from previously collected trajectories. ReasoningBank \cite{9} abstracts successful and failed experiences into generalizable reasoning strategies, while PRINCIPLES \cite{11} generates synthetic strategy memory through offline self-play. MemP \cite{12} supports both offline memory construction and online updates, distilling trajectories into procedural instructions and higher-level abstractions. Other works explore experience-driven graph memory \cite{13,23} and graph-structured skill composition \cite{14}.

Despite advances in experience memory, extracting reliable and transferable correction knowledge from historical trajectories remains challenging. Whether through test-time reflection or offline memory construction, existing methods largely rely on LLMs to diagnose failures and formulate corrective strategies. In contrast, our method uses graph matching to identify common workflows and derive graph edit paths, providing explicit structural guidance for extracting reusable correction knowledge.




\subsection{Graph for Agent Memory}
Recent studies have explored graph memory mechanisms to enhance long-term reasoning and experience reuse in LLM agents \cite{miss43}. AriGraph \cite{miss40} constructs a unified graph integrating semantic and episodic memories for planning and reasoning. A-Mem \cite{35} enables autonomous memory evolution by dynamically linking and restructuring memories into an interconnected network, while G-Memory \cite{miss42} extends graph-based memory to multi-agent settings via hierarchical interaction and insight graphs. More recent efforts, such as SGMem \cite{miss44}, GAM \cite{miss45}, and H-Mem \cite{miss46}, investigate sentence graphs, hierarchical graph memories, and hybrid memory evolution to enhance long-horizon retrieval and adaptive reasoning. In parallel,ExpGraph \cite{miss47} lets frozen executors reuse experience through external graph memory without parameter updates; EXG \cite{miss48} organizes successes and failures into a structured experience graph for self-evolving agents.

Despite encouraging progress, existing methods mainly focus on memory storage and retrieval, while the extraction of reusable high-level insights and the modeling of transferable experience relationships remain underexplored.
Our \model~ fills this gap by using graph matching to derive common subgraphs and edit paths, transforming raw trajectories into transferable correction workflows.

\section{Preliminary}

We consider LLM-based agents that interact with environments to solve long-horizon tasks. Following prior work, we formulate such agentic tasks as a partially observable Markov decision process (POMDP) without explicitly modeling the latent state. The process is defined by the tuple $(\mathcal{Q}, \mathcal{A}, \mathcal{O}, T)$, where $\mathcal{Q}$ is the query (instruction) space, $\mathcal{A}$ the action space, $\mathcal{O}$ the observation space, and $T: \mathcal{A} \times \mathcal{O} \rightarrow \mathcal{A}$ a function that determines the next action given the current action and observation. Since our focus is on the task planning capability of LLM agents, $\mathcal{Q}$, $\mathcal{A}$, and $\mathcal{O}$ are subsets of the natural language space.

Given a task query $q \in \mathcal{Q}$, the LLM agent $\pi_\theta$ interacts with the environment as follows. At each time step $t$, the agent samples an action $a_t \sim \pi_\theta(\cdot \mid q, \tau_{t-1})$ based on the query and the history of previous interactions. It then receives environmental feedback as an observation $o_t \in \mathcal{O}$. The historical interaction trajectory up to step $t-1$ is denoted by $\tau_{t-1} = (o_0, a_1, o_1, a_2, o_2, \ldots, a_{t-1}, o_{t-1})$, where $o_0$ is the initial observation of the environment. The interaction loop continues until the agent either completes the task or reaches the maximum allowed number of steps. The complete trajectory is written as $\tau = (q,, o_0,, a_1,, o_1,, a_2,, o_2,, \ldots,, a_m,, o_m)$, where $o_0$ is the initial observation of the environment, and $m$ denotes the trajectory length. Finally, a trajectory-level reward $r(q, \tau) \in [0,1]$ is assigned to indicate the degree of task completion, where $r(q,\tau)=1$ indicates success, $r(q,\tau)=0$ indicates failure, and intermediate values represent partial success.

\begin{figure*}
    \centering
    \includegraphics[width=\linewidth]{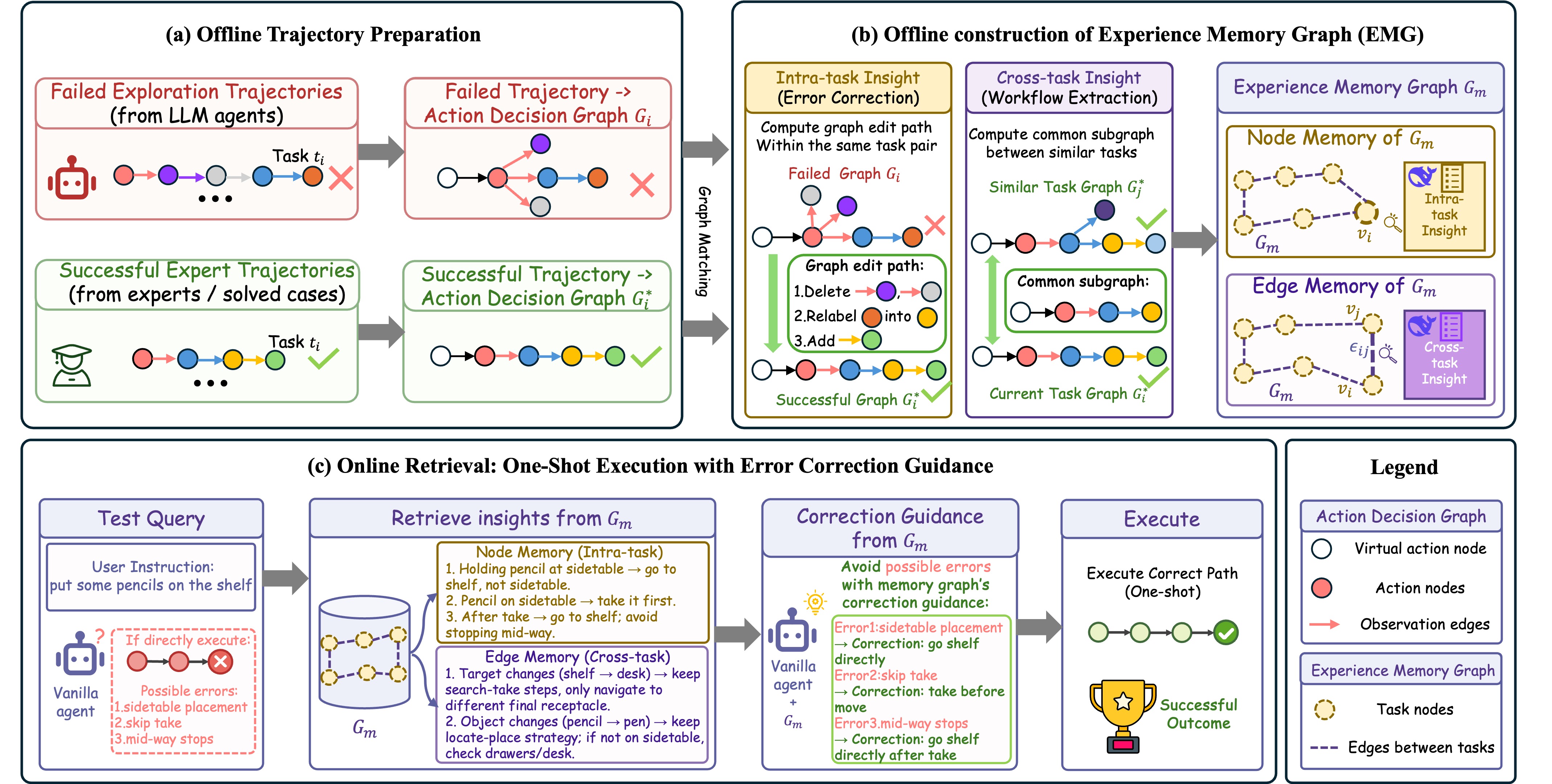}
    \caption{Overall architecture of \model~(Experience Memory Graph). In the offline construction phase, (a) we collect exploration trajectories on training tasks and convert both exploration and provided expert trajectories into action decision graphs. (b) Graph matching on failed–successful pairs extracts edit paths for error correction, while matching on successful–successful pairs from similar tasks yields common subgraphs for transferable workflows. These structural insights are distilled into natural‑language guidance and stored in a memory graph with intra‑task nodes and cross‑task edges. (c) At test time, the agent retrieves relevant node and edge insights, enabling one‑shot execution without trial‑and‑error loops. }
    \label{method:framework}
\end{figure*}

\section{Method}

In this section, we present Experience Memory Graph (\model), a framework that constructs structured experience memory from training tasks and applies it to new tasks, which is shown in fig. 2. The overall pipeline consists of three stages: (a) preparing training trajectories and converting them into action decision graphs; (b) constructing a memory graph where nodes store intra-task insights and edges store cross-task insights; and (c) retrieving relevant memory at test time to guide the agent without additional loops, which corresponds to section 4.1, 4.2 4.3, respectively.

\subsection{Preparation}

Before constructing the experience memory graph, we first collect trajectories on the training tasks and convert them into action decision graphs.

\subsubsection{Collecting Exploration Trajectories}

For each training task query \(q\) in the training set \(\mathcal{Q}_{\text{train}}\), we let an LLM agent interact with the environment to produce a single exploration trajectory \(\tau_i\) (which may either succeed or fail). For the same task \(t_i\),  a successful expert trajectory \(\tau_i^*\) is readily available from the supervised fine-tuning dataset of ETO~\cite{17}, which is a standard component in modern LLM agent pipelines. Thus, for each training task, we obtain a pair \((\tau_i, \tau_i^*)\). Failed exploration trajectories are particularly valuable, as they contain mistakes that can be corrected; successful ones can be used as positive examples directly. We collect such pairs over all training tasks. 

\subsubsection{Transforming Trajectories into Action Decision Graphs}

To explicitly model the decision process — \textit{which action should be taken under a given observation} — we convert each raw trajectory into a directed \textbf{action decision graph}. A successful trajectory and a failed one often differ precisely in this decision mapping. By constructing a graph where nodes are actions and edges carry the preceding observation, we make explicit the rule: \textit{for a given state, which action is chosen}. This decision‑centric representation enables the graph edit path to learn meaningful correction insights — e.g., under a specific observation, which wrong action should be replaced by which correct action. Compared to raw sequential sequences, where the same action may appear in different contexts without a clear link to the triggering observation, our graph formulation yields far more actionable and transferable correction knowledge.

\noindent\textbf{Graph representation.}
We denote the action decision graph as \(G = (V, E, q, r)\), where \(q\) is the task query and \(r\) is the final reward. Nodes represent actions, edges represent the observations preceding those actions. To obtain compact yet discriminative representations for labeling, we normalize each action into a tuple \(k = (\text{type}, \text{object}, \text{receptacle})\) (e.g., \((\text{take}, \text{apple}, \text{table})\)), capturing the action type and the objects it operates on; the raw action text \(a_v^{\text{raw}}\) (e.g., "take apple from table") is also kept. Each node is represented as \(v_i = (k_i, a_i^{\text{raw}}, \ell_n(v))\) with node label \(\ell_n(v_i)\in\mathbb{N}\). A virtual INIT node \(v_{\text{init}}\) (\(\ell_n(v_{\text{init}})=0\)) provides an incoming edge for the first action, carrying the initial observation \(o_0\).

Each directed edge is denoted as \(e_i = ((v_i, v_{i+1}), s_i, \ell_e(e_i)) \in E\). \( (v_i, v_{i+1})\) represents the connection from action node $v_i$ to $v_{i+1}$. So $e_i$ is both the observations resulting from the action $v_i$ and the preceding observation of the next action $v_{i+1}$. $e_i$ also carries a normalized observation text \(s_i\) derived from the raw observation $o_i$ (e.g., “kitchen: apple, potato”) and its edge label \(\ell_e(e_i) \in \mathbb{N}\). 

\noindent\textbf{Construction design.}
Two design choices ensure the action decision graph captures decision-level information. \textbf{1. node reuse}: if the same normalized action tuple appears multiple times in the trajectory, we do not create a new node; instead, we reuse the existing node. This gives the node multiple incoming edges, each associated with a different preceding observation, thus capturing the decision pattern that the same action may be taken under various observed states. \textbf{2. parallelizing consecutive invalid actions}: when an action leads to an uninformative observation (e.g., “nothing happens”), the environment state does not change. We keep the current valid state unchanged and connect all such invalid actions from the same previous valid node, treating them as parallel attempts from the same decision point rather than a spurious chain. 

To implement these ideas, we scan the trajectory once, maintaining the current valid state \(s\) (initialized as normalized \(o_0\)) and the node \(last\) (initially \(v_{\text{init}}\)) that produced it. A dictionary \(\Phi\) maps each normalized action tuple to its node, enabling node reuse. The detailed Algorithm~\ref{alg:build_graph} and a concrete example is available in Appendix A. After building the structure, we assign labels: each distinct action tuple \(k\) receives a unique positive integer as node label \(\ell_n(v)\) (0 for \(v_{\text{init}}\)), and each distinct normalized observation text \(s\) receives a unique positive integer as edge label \(\ell_e(e)\).

The resulting action decision graph abstracts raw action and observation texts into discrete labels (node labels for action tuples, edge labels for observation texts), enabling efficient structure comparison via label equality in the subsequent graph matching stage. At the same time, each node and edge preserves the original raw texts, which are later used to convert extracted common subgraphs and graph edit paths into natural‑language insights.

\subsection{Construction of Experience Memory Graph}

To move beyond test‑time self‑reflection, we build an offline experience memory graph that stores structured, reusable experience. This memory graph provides two key benefits: (1) it captures precise error corrections and successful workflows for individual tasks, and (2) it generalizes across similar tasks by extracting shared decision patterns. To construct such a memory graph, we leverage graph matching on the action decision graphs obtained from Section~4.1. By matching failed exploration graphs with their corresponding expert graphs, we extract intra‑task common subgraphs and graph edit paths; by matching expert graphs of semantically similar tasks, we obtain cross‑task insights. These structured pieces of knowledge are organized into the experience memory graph \(\mathcal{G}_m=(\mathcal{V},\mathcal{E})\), where each node \(\nu \in\mathcal{V}\) stores intra‑task insights and each edge \(\epsilon \in\mathcal{E}\) stores cross‑task reusable insights. We first define the experience memory graph formally.

Let \(\mathcal{V} = \{\nu_1,\dots,\nu_N\}\) where each node \(\nu_i\) is a tuple:
\begin{equation}
\nu_i = \bigl(q_i,\;  {embed}_i,\; I_i^{\text{intra}},\; G_i,\; G_i^*,\;r_i\; \bigr),
\label{eq:memory_node}
\end{equation}
where \(q_i\) is the task query, \({embed}_i\) is the query embedding via nomic-embed-text, \(I_i^{\text{intra}}\) is the intra‑task insight, and \(G_i, G_i^*\) are the action decision graphs of the exploration trajectory and the expert trajectory, respectively (built via Algorithm~\ref{alg:build_graph}). \(r_i\in\{0,1\}\) is the reward (success/failure) of exploration trajectory \(G_i\), and the reward of expert trajectory is always 1 and we omit it here.

Each edge \(\epsilon_{ij}\in\mathcal{E}\) connects \(\nu_i\) and \(\nu_j\) and is defined as:
\begin{equation}
\epsilon_{ij} = \bigl(s_{ij},\; I_{ij}^{\text{cross}}\bigr),
\label{eq:memory_edge}
\end{equation}
where \(s_{ij} = \cos({embed}_i,{embed}_j)\) is the cosine similarity between query embeddings, and \(I_{ij}^{\text{cross}}\) is the cross‑task insight text.

\subsubsection{Constructing Node Insights}

Recall that \(G_i\) and \(G_i^*\) are the action decision graphs built from the raw exploration and expert trajectories using the method in Section~4.1. This graph representation allows us to compare them structurally via graph matching. Given a pair of exploration graph \(G_i\) and expert graph \(G_i^*\), we can extract two kinds of structured knowledge: the \textbf{common subgraph} \(G_i^c\) (the correct actions in the successful/failed trajectory that should be maintained, revealing the successful workflow) and the \textbf{graph edit path} \(G_i^e\) (a sequence of edit operations that transforms failed exploration \(G_i\) into successful expert \(G_i^*\)), from \(G_i^e\) we directly obtain: (a) which actions to add or avoid under specific observations (from node deletions/insertions), and (b) which action to replace with another under a given observation (from relabeled actions). To compute \(G_i^c\) and \(G_i^e\), we first find a node matching relationship between the two graphs via \textbf{graph matching}, then we compare the two graphs: the parts that are identical in both structure and labels form the common subgraph; the differences (mismatched nodes and edges) directly yield the graph edit path. Thus, instead of test‑time trial‑and‑error, we directly obtain correction instructions and successful workflows.

Concretely, we first obtain the node matching between \(G_i\) and \(G_i^*\) via graph matching. Since supervised matching heavily relies on ground‑truth node correspondences, we adopt an unsupervised approach based on optimal transport (OT).  Most OT‑based methods adopt the Fused Gromov‑Wasserstein (FGW) distance \cite{26,29}. Intuitively, this objective balances node relabeling costs and edge structure discrepancies to find a globally coherent node correspondence. For two graphs \(G\) and \(G'\) with soft assignment \(\pi \in \mathbb{R}^{|V| \times |V'|}\), our FGW objective is:
\begin{equation}
\mathrm{FGW} = \min_{\pi} \sum_{i,j} C_{ij}\pi_{ij} + \sum_{i,j,k,l} |A_{ij}^g - A_{kl}^{g'}|^2 \pi_{ik}\pi_{jl},
\label{eq:fgw}
\end{equation}
where \(\pi_{ij}\) indicates the matching strength between node \(i\) in \(G\) and node \(j\) in \(G'\), \(C_{ij}\) is the cost of relabeling node \(i\) to \(j\), and \(A^g, A^{g'}\) are adjacency matrices. Unlike the standard FGW formulation which includes a factor \(\frac{1}{2}\) for undirected graphs where each edge is counted twice) \cite{26,29}, our directed graphs require no such factor, and we drop it. We adopt the FGWAlign solver \cite{26} to optimize this objective, which supports edge-labeled graphs. The time complexity of the solver is \(O(|\mathcal{V}|^3 \cdot |\mathcal{L}|)\), where \(|\mathcal{L}|\) is the number of edge label types. In practice, even for  complex tasks in our benchmarks (which require several dozen action steps), the solver completes graph matching within a few seconds per pair on CPU. Detailed runtime measurements are provided in Section~\ref{sec:graph matching time}.

After obtaining the optimal soft plan \(\pi^*\), we discretize it into a hard node matching matrix \(P \in \{0,1\}^{|V| \times |V'|}\) via linear assignment:
\begin{equation}
P = \arg\max_{P \in \mathbb{P}_n} \langle \pi^*, P \rangle,
\label{eq:linear_assignment}
\end{equation}
where \(\mathbb{P}_n\) denotes the set of permutation (or partial permutation) matrices. Each entry \(P_{ij}=1\) indicates that node \(i\) in \(G\) is matched to node \(j\) in \(G'\).

Given the node matching matrix \(P\), we define a node matching \(M: V \mapsto V'\) where \(M(u)=v\) if \(P_{uv}=1\) (and \(u\) is unmatched if no such \(v\) exists), then we derive the graph edit path and the common subgraph by comparing the two graphs. Classical methods for extracting edit paths and common subgraphs are designed for undirected graphs with only node labels; they cannot handle edge labels and do not respect edge directions. Directly applying them to our directed edge‑labeled action decision graphs would discard the semantic information carried by edge observations (the preconditions of actions). Therefore, we adapt these methods to our setting.  Specifically, Algorithm~\ref{alg:extract_common} retains edges that exist in both graphs with identical labels under the matching (the correct workflow); Algorithm~\ref{alg:derive_edit} collects unmatched or mismatched nodes and edges to form edit operations (deletions, insertions, relabelings), directly indicating the necessary corrections. The detailed algorithm is available in Appendix B.

Finally, we generate the intra‑task insight. If the exploration succeeded ($r_i=1$), we retain the common subgraph $G_i^c$ (extracted via Algorithm~\ref{alg:extract_common}) as the correct workflow. If it failed ($r_i=0$), we use the graph edit path $G_i^e$ (via Algorithm~\ref{alg:derive_edit}) as the correction instructions. For environments with continuous reward (e.g., ScienceWorld), a partially successful trajectory ($0<r_i<1$) contains both correct and incorrect actions; we therefore feed both $G_i^c$ (to preserve the correct actions) and $G_i^e$ (to correct the errors) to the LLM. In all cases, the LLM takes the task query $q_i$ as input, the original exploration graph $G_i$, the expert graph $G_i^*$, and the extracted structure(s) $X_i$ (where $X_i$ is $G_i^c$ ($r_i=1$), or $G_i^e$ ($r_i=0$), or both ($0<r_i<1$)), and summarizes them into intra-task insight:
\begin{equation}
I_i^{\text{intra}} = \mathrm{LLM}\bigl( \text{intra-prompt}(q_i, G_i, G_i^*, X_i) \bigr).
\label{eq:intra_insight}
\end{equation}

In summary, \(I_i^{\text{intra}}\) supplies either the correct workflow or explicit corrections, enabling test‑time guidance without trial‑and‑error.

\subsubsection{Constructing Edge Insights}

Using the action decision graph representation for all expert trajectories, we can compare the successful decision patterns of different tasks. We further aim to extract generalizable knowledge that transfers across similar tasks (e.g., different environments with analogous goals). For two task nodes \(\nu_i\) and \(\nu_j\) with similar queries in the memory graph, comparing their expert graphs \(G_i^*\) and \(G_j^*\) via graph matching reveals a common subgraph — actions that are correct under similar observations in different environments which should be learned and maintained. This provides cross‑task insights that are robust to environmental variations.

To realize this, after building all nodes we compute pairwise query embedding similarities \(s_{ij} = \cos({embed}_i, {embed}_j)\), where \({embed}_i\) is the embedding of node \(\nu_i\)'s query. An edge \(\epsilon_{ij}\) is added if \(s_{ij} \ge \tau_{\epsilon}\) (we set \(\tau_{\epsilon}=0.8\)), and we only connect each node to its top-\(k\) most similar neighbors. For each edge \(\epsilon_{ij}\) connecting node task \(\nu_i\) and \(\nu_j\), we perform graph matching between the expert graphs \(G_i^*\) and \(G_j^*\) of \(\nu_i\) and \(\nu_j\), and use the same adapted FGW solver and Algorithm~\ref{alg:extract_common} to obtain the common subgraph \(G_{ij}^c\). We then prompt an LLM to summarize \(G_{ij}^c\) into the cross‑task insight:
\begin{equation}
I_{ij}^{\text{cross}} = \mathrm{LLM}\bigl( \text{cross\_prompt}(q_i, q_j, G_i^*, G_j^*, G_{ij}^c) \bigr).
\label{eq:cross_insight}
\end{equation}
This insight is stored on the edge $\epsilon_{ij}$, which can guide the agent when facing new tasks similar to either endpoint.

\subsection{Retrieval of Experience Memory Graph}

During test time, given a new task query \(q_{\text{test}}\), we first embed it using the same embedding model to obtain \({embed}_{\text{test}}\). The similarity to each memory node \(\nu_i\) is computed as \(s_i = \cos({embed}_{\text{test}}, {embed}_i)\). We retrieve the top-\(K\) nodes with the highest \(s_i\). For each such node \(\nu_i\), we also retrieve the most relevant cross‑task insights: we look at all edges incident to \(\nu_i\), and for each neighbor \(\nu_j\) we compute its similarity to \(q_{\text{test}}\) as above. The edges with the top-\(T\) neighbor similarities are selected, and their stored cross‑task insights are collected. Formally, the retrieval result for \(q_{\text{test}}\) is:
\begin{equation}
\mathcal{R}(q_{\text{test}}) = \bigcup_{\nu_i \in \text{TopK}} \left\{ \bigl( I_i^{\text{intra}},\; \{ I_{ij}^{\text{cross}} \mid \nu_j \in \text{TopT}(\nu_i) \} \bigr) \right\},
\label{eq:retrieval}
\end{equation}
where \(\text{TopT}(\nu_i)\) are the neighbors of \(\nu_i\) with the highest query similarities.
The retrieved insights, derived from structured graph comparisons, provide concrete decision rules, making them directly usable for the agent. For instance, when faced with a task such as “put some pencil on shelf,” the retrieved node insight may specify that once the agent is holding the pencil at the sidetable, the correct next action is to go directly to the shelf rather than placing it elsewhere, and that the action of taking the pencil must precede any movement toward the target. The retrieved edge insights further provide cross‑task guidance, such as adapting the final navigation step when the target changes from shelf to desk, or expanding the search strategy when the object changes from pencil to pen. These insights are then passed to the agent’s prompt as guidance, eliminating any need for test‑time reflection or trial‑and‑error loops.

The complete Experience Memory Graph \(\mathcal{G}_m\) is built offline on the training set. Its construction cost is amortized over many test tasks, and the retrieval is fast (embedding similarity + neighbor lookup). This makes \model\  both efficient and scalable.

\section{Experiments}
\subsection{Experimental Settings}
\begin{table}[h]
\centering
\caption{Statistics of the datasets used for evaluation. }
\label{tab:datasets}
\resizebox{1.0\linewidth}{!}{
\begin{tabular}{lcccc}
\toprule
\textbf{Dataset} & \textbf{\#Train} & \textbf{\#Test-Seen} & \textbf{\#Test-Unseen}  \\
\midrule
ALFWorld     & 3321 & 140 & 134  \\
ScienceWorld & 1483 & 194 & 211  \\

\bottomrule
\end{tabular}
}
\end{table}
\subsubsection{Datasets}
 We conduct experiments across two  mainstream datasets: ALFWorld and ScienceWorld. The detailed statistics are summarized in Table~\ref{tab:datasets}. ALFWorld features embodied household tasks, while ScienceWorld simulates interactive science experiments with longer trajectories and more complex multi-step reasoning. We evaluate both in-distribution and out-of-distribution generalization on seen and unseen test splits, following the same ETO~\cite{17} split. ALFWorld provides binary rewards (0 or 1), whereas ScienceWorld uses continuous rewards in $[0,1]$.

\subsubsection{Baselines}
We benchmark \model~ against representative agent frameworks, covering both test-time reflection methods and memory construction methods. We first consider ReAct~\cite{32}, a widely adopted prompting strategy that interleaves reasoning traces with task-specific actions, serving as our fundamental memory-less baseline.
\textbf{Test-time reflection:} We evaluate Reflexion~\cite{33}, which uses verbal feedback and episodic memory to iteratively refine agent behavior; ExpeL~\cite{34}, which extracts generalizable insights from past trajectories to improve subsequent task execution; and CDMem~\cite{15}, which combines short-term and long-term memory with reflection to support context-aware experience reuse. For these methods, we allow five reflection iterations at test time.
\textbf{Memory construction:} We further compare against MemP~\cite{12}, which distills trajectories into procedural instructions and higher-level abstractions; ReasoningBank~\cite{9}, which abstracts successful and failed experiences into reusable reasoning strategies; and A-Mem~\cite{35}, which dynamically organizes and links memories into an interconnected network to support adaptive memory retrieval and evolution.
Our \model~ uses ReAct as the base agent and retrieves insights from the constructed experience memory graph at test time. All methods are equipped with few-shot examples to ensure a consistent evaluation setting.

\subsubsection{Evaluation Metrics and Settings}
We use Success Rate (SR), the proportion of tasks completed successfully, and Average Reward (AR), the mean reward across all tasks, to evaluate task completion and partial progress, respectively. For EMG, we consistently use DeepSeek V3.2 to collect exploration trajectories. For testing, we use both the open-source small model Qwen3:4B-instruct-2507-q8 and the large model DeepSeek-V4-Flash to evaluate the baselines. For memory construction baselines, we use SFT trajectories provided by ETO, including 1,275 trajectories whose tasks overlap with the ALFWorld training set and 1,483 trajectories for ScienceWorld, to construct memory offline using the same LLM as for test-time task execution.

\begin{table}[htbp]
\centering
\caption{Main results of our \model~compared with various baselines, where DeepSeek is abbreviated in DS. S-SR and U-SR denote Success Rate (\%) on Seen and Unseen splits respectively, and S-AR and U-AR denote Average Reward (\%) on Seen and Unseen splits.}
\label{tab:main_results}
\resizebox{1.0\linewidth}{!}{
\begin{tabular}{ll|cc|cccc|c}
\toprule
& \multirow{2}{*}{\textbf{Method}} & \multicolumn{2}{c|}{\textbf{ALFWorld}} & \multicolumn{4}{c|}{\textbf{ScienceWorld}} & \multirow{2}{*}{\textbf{Avg.}} \\
\cmidrule(lr){3-4} \cmidrule(lr){5-8}
& & \textbf{S-SR} & \textbf{U-SR} & \textbf{S-AR} & \textbf{S-SR} & \textbf{U-AR} & \textbf{U-SR} & \\
\midrule
\multirow{6}{*}{\rotatebox{90}{\textit{Qwen3-4B}}} 
& ReAct     & 29.29 & 21.64 & 33.19 & 11.86 & 30.57 & 8.53 & 22.51 \\
& Reflexion & 34.29 & 39.55 & 39.25 & \second{14.95} & 30.06 & 8.06 & 27.69 \\
& ExpeL     & \second{38.57} & 39.55 & 39.94 & 13.40 & 35.69 & 11.85 & 29.83 \\
& CDMem     & 27.14 & 32.84 & 29.99 & 13.92 & 23.13 & 8.53 & 22.59 \\
& MemP      & 37.86 & \second{42.54} & 26.83 & 9.28 & 25.20 & 8.53 & 25.04 \\
& Reasoning Bank  &23.57 &22.29 &37.33 &11.34 &33.78 &8.53 & 22.81  \\
& A-Mem  &\second{38.57} &33.58 &\second{41.79} &14.43 &\second{42.80} &\second{15.17} &\second{31.06}   \\
& EMG       & \best{53.57} & \best{51.49} & \best{50.69} & \best{18.04} & \best{48.21} & \best{15.64} & \best{39.61} \\
\midrule
\multirow{6}{*}{\rotatebox{90}{\textit{DS-V4-Flash}}}
& ReAct     & 86.43 & 88.06 & 51.77 & 17.53 & 52.69 & 13.74 & 51.70 \\
& Reflexion & 82.86 & \second{90.30} & 47.12 & 12.89 & 43.92 & 8.06 & 47.53 \\
& ExpeL     & 81.43 & 83.58 & 46.56 & 19.07 & 44.90 & 15.17 & 48.45 \\
& CDMem     & \second{95.71} & \second{90.30} & 31.91 & 15.46 & 35.38 & 8.53 & 46.22 \\
& MemP      & 85.71 & 85.82 & 44.73 & 10.31 & 44.78 & 11.37 & 47.12 \\
& Reasoning Bank  &81.43 &88.81 &51.47 &\second{22.16} &53.83 &18.01 &52.62   \\
& A-Mem  &92.14 &89.55 &\second{57.84} &20.10 &\second{57.72} &\second{17.54} &\second{55.82}   \\
& EMG       & \best{96.43} & \best{96.27} & \best{60.20} & \best{31.44} & \best{61.84} & \best{26.54} & \best{62.12} \\
\bottomrule
\end{tabular}
}
\end{table} 
\subsection{Main Results}
The main results are presented in Table~\ref{tab:main_results}, from which we draw three observations.

\textbf{Consistent Performance Improvements.} EMG achieves the highest average score among all evaluated methods with both Qwen3-4B and DeepSeek-V4-Flash. It also consistently outperforms the baselines on both ALFWorld splits and all four ScienceWorld metrics. These results suggest that explicitly extracting correction paths and reusable workflows can complement conventional reflection and memory construction methods.

\textbf{Greater Benefits for Smaller Models.} EMG yields larger relative improvements over the strongest baseline with Qwen3-4B than with DeepSeek-V4-Flash, particularly on ALFWorld. This suggests that structured experience memory can help smaller models make better decisions by providing reusable corrective insights, potentially reducing their reliance on self-reflection.

\textbf{Improved Performance on Complex Tasks.} On ScienceWorld, EMG achieves the best results across both seen and unseen splits with both models. Notably, Qwen3-4B equipped with EMG achieves performance comparable to several DeepSeek-V4-Flash-based baselines, suggesting that EMG can help smaller models tackle complex, long-horizon tasks requiring multi-step reasoning.

\begin{table}[htbp]
\centering
\caption{Ablation study of \model~.}
\label{tab:ablation}
\resizebox{1.0\linewidth}{!}{
\begin{tabular}{ll|cc|cccc}
\toprule
& \multirow{2}{*}{\textbf{Method}} & \multicolumn{2}{c|}{\textbf{ALFWorld}} & \multicolumn{4}{c}{\textbf{ScienceWorld}} \\
\cmidrule(lr){3-4} \cmidrule(lr){5-8}
& & \textbf{S-SR} & \textbf{U-SR} & \textbf{S-AR} & \textbf{S-SR} & \textbf{U-AR} & \textbf{U-SR} \\
\midrule
\multirow{3}{*}{\rotatebox{90}{\textit{Qwen3}}}
& \model~w/o NE & 29.29 & 21.64 & 33.19 & 11.86 & 30.57 & 8.53 \\
& \model~w/o E &40.71  &38.81  &40.52  &15.98  &38.38  &12.80  \\
& \model~w/o GM &21.43  &31.34  &43.97  &15.46  &40.31  &10.90  \\
& \model &53.57  &51.49  &50.69  &18.04  &48.21  &15.64  \\
\midrule
\multirow{3}{*}{\rotatebox{90}{\textit{DS-V4}}}
& \model~w/o NE & 86.43 & 88.06 & 51.77 & 17.53 & 52.69 & 13.74 \\
& \model~w/o E &90.00  &91.79  &58.91  &27.84  &60.15  &24.17  \\
& \model~w/o GM &82.14  &86.57  &49.22  &14.43  &53.05  &14.22  \\
& \model &96.43  &96.27  &60.20  &31.44  &61.84  &26.54 \\
\bottomrule
\end{tabular}
}
\end{table}
\subsection{Ablation Study}
To isolate the contributions of each component, we evaluate three ablated variants: 
 \textbf{\model~w/o NE}: removing both node's intra-task insights and edge's cross-task insights, and it is actually ReAct without memory;
 \textbf{\model~w/o E}: removing only the edge's cross-task insights and retaining node's intra-task insights.
 \textbf{\model~w/o GM}: removing graph edit path and common subgraph and only prompting an LLM to reflect on errors and extract successful workflows from paired success–failure trajectories to construct memory.
The results are detailed in Table~\ref{tab:ablation}, from which we draw three principal conclusions.

\textbf{Importance of Intra-Task Node Memory.} Comparing \model~w/o NE with \model~w/o E shows that intra-task insights provide substantial improvements across most settings. By preserving task-specific corrective knowledge, node memory helps agents avoid repeated errors and execute tasks more effectively, although its benefits vary across environments and model sizes.

\textbf{Benefits of Cross-Task Insights.} Comparing \model~w/o E with the full \model~ shows that incorporating cross-task insights further improves performance across both datasets and model sizes. These results suggest that edge-based knowledge transfer complements task-specific memory, particularly when agents encounter tasks requiring the reuse of experience from related tasks.

\textbf{Effectiveness of Graph Matching.} Comparing the full \model~ with \model~w/o GM demonstrates the value of our graph-matching-based extraction mechanism. Given the same paired successful and failed trajectories, simply prompting an LLM to reflect on failures and extract successful workflows is less effective than explicitly identifying common subgraphs and graph edit paths. This suggests that graph matching provides a more effective means of extracting reusable workflows and corrective knowledge from historical experiences.

\begin{table}[htbp]
\centering
\caption{Influence of LLM's size for \model~ on ScienceWorld. Qwen and DeepSeek represent that the memory is constructed via Qwen3-4B and DeepSeek-V4 Flash, respectively.}
\label{tab:LLM-influence}
\resizebox{0.7\linewidth}{!}{
\begin{tabular}{l|cccc}
\toprule
\textbf{Method} & \textbf{S-AR} & \textbf{S-SR} & \textbf{U-AR} & \textbf{U-SR} \\
\midrule
Qwen & 50.69 & 18.04 & 48.21 & 15.64 \\
Deepseek & 52.16 & 18.56 & 48.00 & 14.22 \\
\bottomrule
\end{tabular}
}
\end{table}

\subsection{Influence of LLM’s Size for EMG}

We investigate the influence of the LLM's size used to summarize graph edit paths and common subgraphs into node and edge memories on the challenging ScienceWorld benchmark. Specifically, we compare Qwen3-4B and DeepSeek-V4 Flash for memory construction, while keeping Qwen3-4B as the test-time execution model for both settings.

As shown in Table~\ref{tab:LLM-influence}, the two configurations achieve comparable performance across all metrics. This suggests that EMG’s graph edit paths and common subgraphs provide sufficiently explicit corrective knowledge and reusable workflows, reducing the dependence of memory construction on the LLM’s own reasoning and summarization capabilities. By grounding memory construction in structured trajectory matching, EMG enables even a small model to extract useful memories for complex tasks, further demonstrating the value of our graph-matching-based approach.

\subsection{One‑Shot Memory vs. Iterative Correction}

We evaluate all methods on the seen splits of ALFWorld (with Qwen3-4B) and ScienceWorld (with DeepSeek-V4-Flash). For iterative baselines (Reflexion, ExpeL, CDMem), we allow up to five test-time iterations and report their final success rate. For one-shot memory, they perform a single execution per test task without online trial-and-error. The results are shown in figure 3.

Although iterative reflection can improve performance over simpler baselines, \model~ achieves the highest success rate on both benchmarks with a single execution per task, outperforming methods that apply multiple rounds of test-time reflection. Moreover, compared with other one-shot memory construction methods, \model~ achieves better performance by more effectively extracting corrective knowledge and reusable workflows from historical trajectories. These results highlight the benefits of structured experience extraction for improving task success.

\begin{figure}[t]
    \centering
    \includegraphics[width=\linewidth]{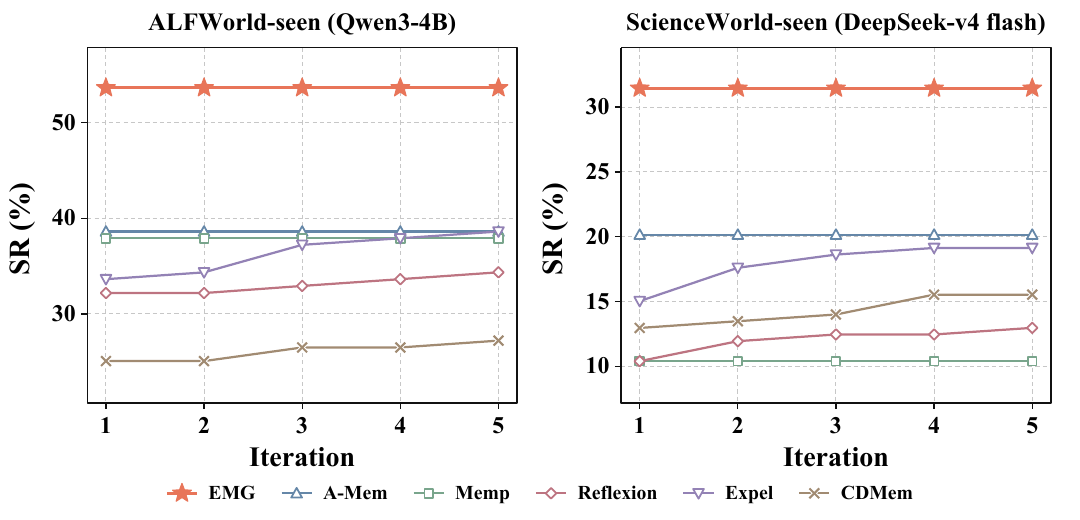}
    \caption{Success rate between one-shot memory and iterative correction methods. The one-shot memory methods (EMG, A-Mem and Memp) only run for 1 iteration, and the iterative methods (Reflexion, Expel and CDMem) run for 5 iterations.}
    \label{intro:sketch}
\end{figure}
\subsection{Hyperparameter Sensitivity Analysis}
We analyze the sensitivity of \model~to the number of retrieved nodes (top‑\(K\)) and retrieved edges (top‑\(T\)) on two benchmarks: ALFWorld unseen and ScienceWorld unseen, both using DeepSeek-V4-Flash. When varying top-\(K\) to test sensitivity, we fix top-\(T\); when varying top-\(T\), we fix top-\(K\), and we report the success rate in figure 4.

Overall, \model~exhibits relatively high robustness across the tested ranges of both the number of retrieved nodes (top-\(K\)) and the number of retrieved edges (top-\(T\)). Performance remains consistently high with only minor fluctuations, and even the smallest retrieval sets achieve competitive results, indicating that the stored insights are highly informative and do not require many examples to be effective. The variation is modest, confirming that \model~is not sensitive to the exact choices of \(K\) and \(T\) and performs reliably without extensive tuning.

\begin{figure}[t]
    \centering
    \includegraphics[width=\linewidth]{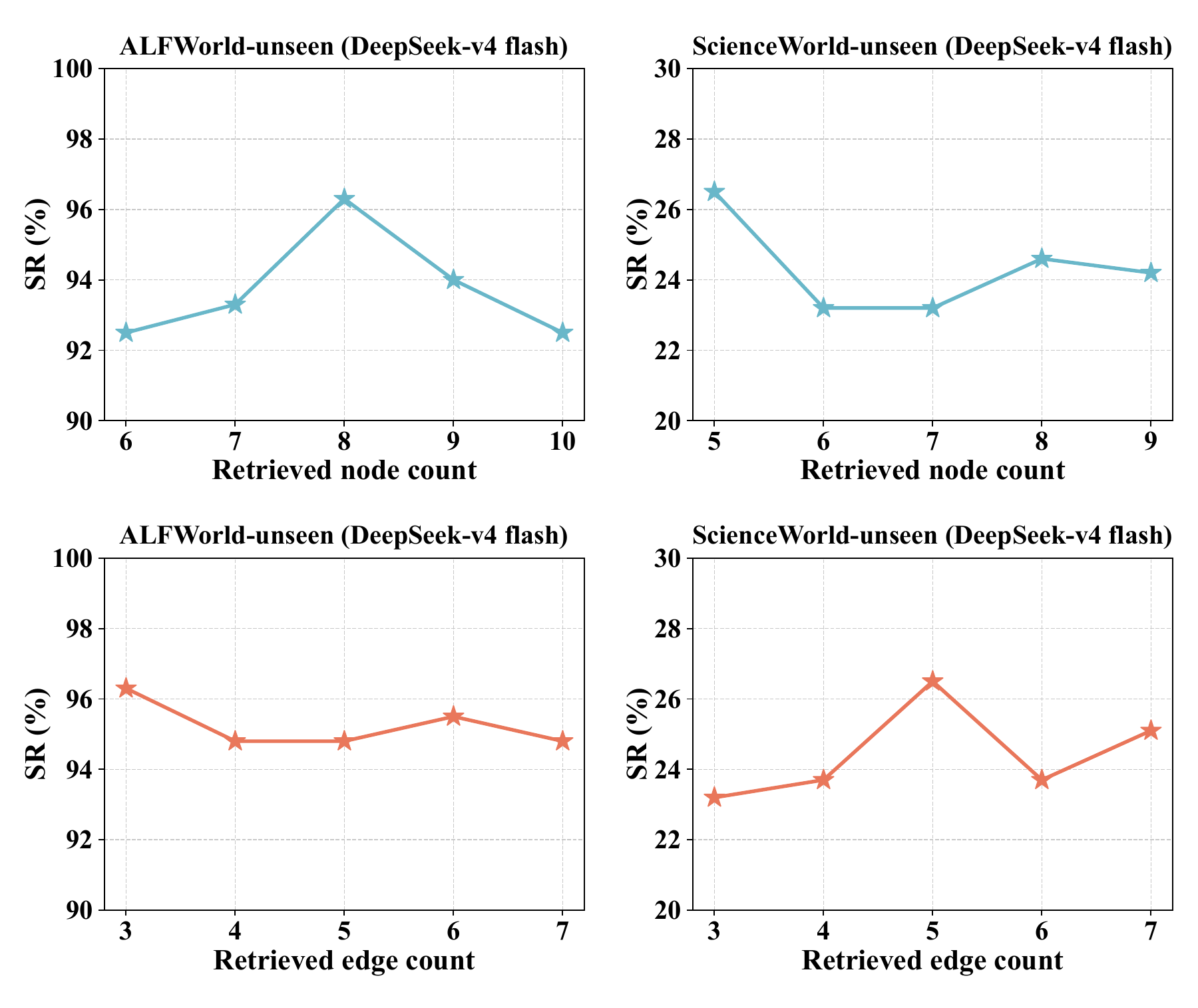}
    \caption{Sensitivity analysis of EMG. We report the success rate of EMG when the retrieved node counts (above) and edge counts (below) change.}
    \label{intro:sketch}
\end{figure}

\begin{figure}[htbp]
    \centering
    \includegraphics[width=1.0\linewidth]{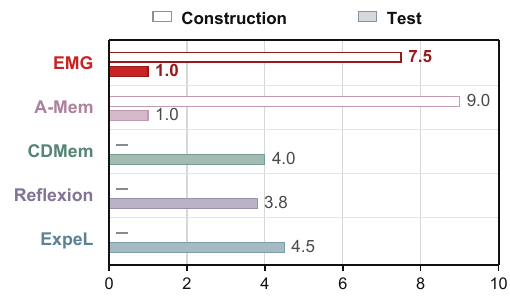}
    \caption{Time cost (hours) of different methods. Construction denotes offline memory construction, while Test denotes the total test-time cost.}
    \label{time cost}
\end{figure}

\subsection{Time Cost Analysis}
\label{sec:graph matching time}
We analyze the overall time cost on ScienceWorld using Qwen3-4B with 4×A800 GPUs. As shown in Fig~\ref{time cost}, we report the offline memory construction time and the test-time cost for EMG, A-Mem, CDMem, Reflexion, and ExpeL. Only EMG and A-Mem require an offline memory construction phase; the other three methods build no persistent memory and thus have zero construction time. For EMG, the construction consists of collecting exploration trajectories (0.5\,h), graph matching and building graph edit path and common subgraph (1.5\,h), and using the LLM to generate node and edge insights (5.5\,h), totaling 7.5\,h.

Although EMG requires an offline construction phase, its cost is comparable to or somewhat lower than A-Mem, since EMG's graph matching and insight generation can be parallelized, whereas methods that incrementally link new entries to an existing memory graph, such as A-Mem, are more sequential. At test time, EMG needs only a single execution per task, while reflection-based methods require multiple rounds, so EMG incurs lower per-task inference cost once the memory is built.



\section{Conclusion}
We propose Experience Memory Graph (EMG), a framework for extracting reusable correction knowledge from agent trajectories through graph matching. Despite advances in experience memory, reliably identifying transferable corrective insights remains challenging. EMG addresses this challenge by converting failed exploration and successful expert trajectories into directed action decision graphs, then matching them to extract common subgraphs that represent successful workflows and graph edit paths that specify how failures can be corrected. These insights form a memory graph, with nodes storing intra-task knowledge and edges capturing cross-task patterns, which agents retrieve at test time. Experiments on ALFWorld and ScienceWorld demonstrate that EMG consistently outperforms strong reflection and memory construction baselines, and ablation studies validate the contributions of graph matching. Future work could explore incorporating self-generated or weakly supervised successful trajectories to broaden EMG's applicability without pre-collected expert demonstrations.


\bibliographystyle{ACM-Reference-Format}
\bibliography{gmagent}

\appendix

\newpage
\appendix
\begin{figure}[ht]
    \centering
    \includegraphics[width=\linewidth]{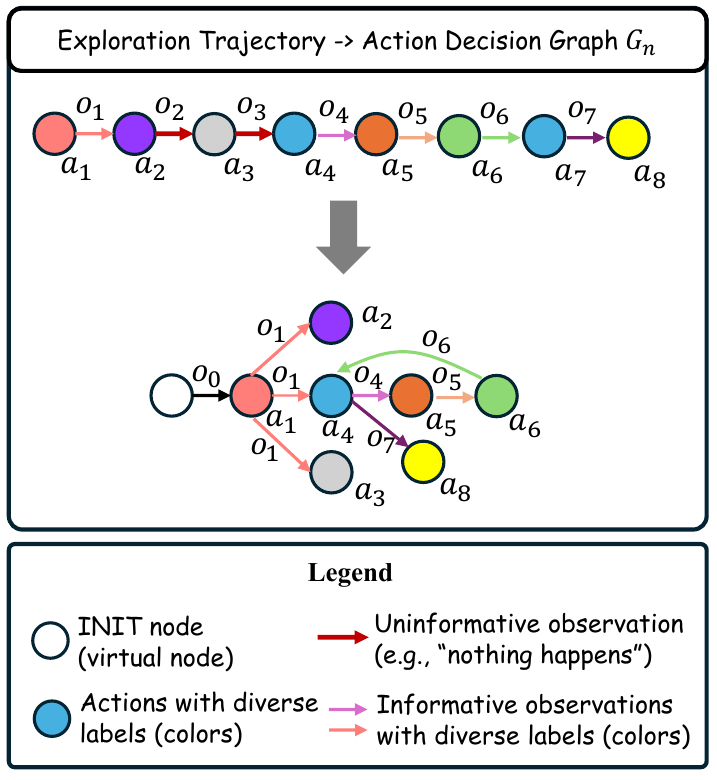}
    \caption{A concrete example for constructing action decision graph. Nodes/ Edges with different colors represent different types of actions/ observations.}
    \label{intro:sketch}
\end{figure}
\section{Details for constructing action decision graph}

Algorithm~\ref{alg:build_graph} details the procedure of constructing the directed action decision graph. Figure 6 demonstrates a concrete example of how to transform the sequential raw exploration trajectory $\tau_n$ into directed action decision graph $G_n$, where $o_i$ represents observations and $a_i$ represents actions. 

In the raw trajectory $\tau_n$, observations $o_2$ and $o_3$ are uninformative (e.g., ``nothing happened'' in ALFWorld), indicating that actions $a_2$ and $a_3$ are both invalid attempts under the same valid observation $o_1$. The subsequent action $a_4$ is a valid action also taken under $o_1$. If we directly applied graph edit distance to the raw sequential trajectory, we would delete $o_2, o_3, a_2, a_3$; however, as $o_2$ and $o_3$ carry no meaningful information, the resulting edit path would produce an insight such as ``under observation $o_2$ do not take $a_3$'', which is meaningless. The correct insight should be ``under observation $o_1$, avoid $a_2$ and $a_3$''. To capture this, we introduce \textbf{parallelizing consecutive invalid actions}: all actions that follow the same valid state $o_1$ without changing the environment are connected in parallel from the same previous valid node. Hence $a_2$, $a_3$, and $a_4$ all receive edges from $a_1$ with the same edge label $o_1$. This makes the graph edit path directly indicate which actions are invalid under $o_1$.

Additionally, we observe that the same core action may be taken under different observations. To better model this, we propose \textbf{node reuse}: if a normalized action appears later in the trajectory (e.g., $a_7$ is the same action as $a_4$), we do not create a new node; instead, we connect the incoming edge (carrying $o_6$) directly to the existing node of that action ($a_4$). This allows a single node to collect multiple incoming edges with different observation labels, revealing under which conditions that action is appropriate.

\begin{algorithm}[tb]
\caption{Building Action Decision Graph}
\label{alg:build_graph}
{\raggedright
\textbf{Input}: Trajectory \(\tau=(q,o_0,a_1,o_1,\dots,a_m,o_m)\) and its reward \(r\).\\}
{\raggedright
\textbf{Output}: Directed labeled multigraph \(G=(V,E,q,r)\), where every edge is an
occurrence \((u,v,\kappa)\) keyed by its step index \(\kappa\); node attributes
\((k,a^{\text{raw}})\), edge attributes (pre-action state \(s\), raw action, observation);
labels \(\ell_n,\ell_e\) assigned post-hoc.}

\begin{algorithmic}[1]
\STATE \(s \leftarrow \text{normalize}(o_0)\); \(last \leftarrow v_{\text{init}}\); \(\Phi \leftarrow \{\}\)
\FOR{\(t = 1\) to \(m\)}
    \STATE \(k \leftarrow \text{normalize}(a_t^{\text{raw}})\)
    \IF{\(k \notin \Phi\)}
        \STATE Create node \(v\) with \((k, a_t^{\text{raw}})\); \(\Phi[k] \leftarrow v\)
    \ENDIF
    \STATE \(v \leftarrow \Phi[k]\)
    \STATE \(\tilde{o} \leftarrow \text{normalize}(o_t)\)
    \STATE Add occurrence edge \((last, v, t)\) with state \(s\), raw action
           \(a_t^{\text{raw}}\), observation \(o_t\)
    \IF{\(\tilde{o}\) is valid \textbf{and} \(\tilde{o}\neq s\)}
        \STATE \(s \leftarrow \tilde{o}\); \(last \leftarrow v\)
    \ENDIF
\ENDFOR
\STATE \(G.q \leftarrow q\); \(G.r \leftarrow r\)
\STATE \textbf{return} \(G\)
\end{algorithmic}
\end{algorithm}

\begin{algorithm}[tb]
\caption{Extract Common Subgraph from Node Matching}
\label{alg:extract_common}
{\raggedright
\textbf{Input}: Directed labeled multigraphs \(G=(V,E,\ell_n,\ell_e)\),
\(G'=(V',E',\ell_n',\ell_e')\), a node matching \(M:V\rightharpoonup V'\) (injective on
matched nodes), the soft assignment \(\pi\), and confidence threshold \(\theta\).\\}
{\raggedright
\textbf{Output}: Common subgraph \(G_c=(V_c,E_c)\), on which the matching preserves both
node labels and edge labels.}

\begin{algorithmic}[1]
\STATE \(V_c \leftarrow \emptyset\); \(M_c \leftarrow \{\}\); \(E_c \leftarrow \emptyset\);
       \(U \leftarrow \emptyset\)
\FOR{each \(u\in V\) with \(v=M(u)\) defined and \(\pi_{uv}\ge\theta\)}
    \IF{\(\ell_n(u) = \ell_n'(v)\)}
        \STATE \(V_c \leftarrow V_c\cup\{u\}\); \(M_c[u] \leftarrow v\)
    \ENDIF
\ENDFOR
\FOR{each occurrence \((u_1,u_2,\kappa)\in E\) in step order with \(u_1,u_2\in V_c\)}
    \STATE \((x,y) \leftarrow (M_c(u_1), M_c(u_2))\)
    \IF{\(\exists\,(x,y,\kappa')\in E'\setminus U\) with
        \(\ell_e'(x,y,\kappa') = \ell_e(u_1,u_2,\kappa)\)}
        \STATE \(E_c \leftarrow E_c\cup\{(u_1,u_2,\kappa)\}\);
               \(U \leftarrow U\cup\{(x,y,\kappa')\}\)
    \ENDIF
\ENDFOR
\STATE \textbf{return} \(G_c\)
\end{algorithmic}
\end{algorithm}

\begin{algorithm}[tb]
\caption{Derive and Apply Graph Edit Path}
\label{alg:derive_edit}
{\raggedright
\textbf{Input}: Same as Algorithm~\ref{alg:extract_common}; \(\mathrm{ep}(\cdot)\) denotes
the set of ordered endpoint pairs of an edge multiset.\\
\textbf{Output}: ID-aligned source \(\tilde G\); node edit sets
\(\Delta_n^{\text{del}}\), \(\Delta_n^{\text{ins}}\), \(\Delta_n^{\text{relab}}\);
edge edit sets \(\Delta_e^{\text{del}}\), \(\Delta_e^{\text{ins}}\),
\(\Delta_e^{\text{relab}}\); and the edited graph \(H\).\par}

\begin{algorithmic}[1]
\STATE \textbf{Phase 1: node-ID alignment}
\FOR{each \(u\in V\)}
    \STATE \(\rho(u) \leftarrow M(u)\) if defined, otherwise a fresh ID \(\notin V'\)
\ENDFOR
\STATE \(\tilde G \leftarrow \rho(G)\) with all labels of \(G\) retained;
       \(V_{\text{com}} \leftarrow \rho(V)\cap V'\)
\STATE \textbf{Phase 2: edit extraction}
\STATE \(\Delta_n^{\text{del}} \leftarrow
       \{(w,\tilde\ell_n(w)) \mid w\in\rho(V)\setminus V'\}\)
\STATE \(\Delta_n^{\text{ins}} \leftarrow
       \{(w,\ell_n'(w)) \mid w\in V'\setminus\rho(V)\}\)
\STATE \(\Delta_n^{\text{relab}} \leftarrow
       \{(w,\tilde\ell_n(w),\ell_n'(w)) \mid w\in V_{\text{com}},\
       \tilde\ell_n(w)\neq\ell_n'(w)\}\)
\FOR{each ordered pair \((x,y)\in \mathrm{ep}(\tilde E)\cup \mathrm{ep}(E')\)}
    \STATE \(S \leftarrow\) occurrences of \((x,y)\) in \(\tilde E\) sorted by step index
    \STATE \(T \leftarrow\) occurrences of \((x,y)\) in \(E'\) sorted by step index
    \IF{\(x\in V_{\text{com}}\) \textbf{and} \(y\in V_{\text{com}}\)}
        \STATE \(c \leftarrow \min(|S|,|T|)\)
    \ELSE
        \STATE \(c \leftarrow 0\)
    \ENDIF
    \STATE \(\Delta_e^{\text{relab}} \mathrel{{+}{=}}
           \{(S_i\!\rightarrow\!T_i)\mid i\le c\}\), keeping a pair only when
           \(S_i\) and \(T_i\) differ in edge label or attributes
    \STATE \(\Delta_e^{\text{del}} \mathrel{{+}{=}} S_{c+1..|S|}\);\quad
           \(\Delta_e^{\text{ins}} \mathrel{{+}{=}} T_{c+1..|T|}\)
\ENDFOR
\STATE \textbf{Phase 3: edit application}
\STATE \(H \leftarrow \tilde G\)
\STATE Remove from \(H\) every occurrence in \(\Delta_e^{\text{del}}\)
\STATE Set \(\ell_n^H(w) \leftarrow b\) for each \((w,a,b)\in\Delta_n^{\text{relab}}\)
\STATE Replace the label and attributes of \(e\) by those of \(e'\) for each
       \((e\rightarrow e')\in\Delta_e^{\text{relab}}\)
\STATE Remove from \(H\) every node in \(\Delta_n^{\text{del}}\), which is isolated
       after line 23
\STATE Add to \(H\) every node in \(\Delta_n^{\text{ins}}\) with its recorded label
\STATE Add to \(H\) every occurrence in \(\Delta_e^{\text{ins}}\) with its recorded label,
       its endpoints being present after line 27
\STATE \textbf{return} \(\tilde G\), \(\Delta_n^{\ast}\), \(\Delta_e^{\ast}\), \(H\)
\end{algorithmic}
\end{algorithm}

\section{Algorithm of extracting common subgraph and graph edit path}

Given a node matching \(M\) between two directed labeled multigraphs \(G\) and \(G'\)
(obtained via the FGW solver), where every edge is an occurrence \((u,v,\kappa)\) keyed by its
step index, the following two algorithms extract structured knowledge from the matched graphs.

\textbf{Algorithm~\ref{alg:extract_common}} constructs the common subgraph. Since the FGW
solver may pair nodes carrying different action labels, it first restricts \(M\) to its
label-consistent part \(M_c\), and then retains an occurrence of \(G\) only if an as yet
unconsumed occurrence with the same edge label exists between the images of its endpoints, so
multiplicity is never over-counted. The matching therefore preserves both node and edge labels
on \(G_c\), which is the successful workflow; a matched pair with different action labels is a
substitution and is handled by Algorithm~\ref{alg:derive_edit} as a node relabeling.

\textbf{Algorithm~\ref{alg:derive_edit}} derives the edit path and applies it. Phase 1 renames
the node IDs of \(G\) by \(M\), sending unmatched nodes to fresh IDs disjoint from \(V'\), so
both graphs share one ID space. Phase 2 collects unmatched nodes and node label mismatches,
and for every ordered endpoint pair occurring in \emph{either} graph pairs the two occurrence
lists by step order, surplus occurrences becoming deletions in \(\tilde G\) and insertions in
\(G'\). Phase 3 applies the operations in a fixed order --- edge deletions, node relabelings,
edge relabelings, node deletions, node insertions, edge insertions --- which is where the edit
semantics are pinned down. Because line 10 ranges over the union of the endpoint pairs and
lines 18--19 partition each bucket, every occurrence is covered exactly once; line 16 forces
\(c=0\) at an unmatched endpoint, so the nodes of \(\Delta_n^{\text{del}}\) are isolated and
the endpoints of \(\Delta_e^{\text{ins}}\) exist when used, giving \(H=G'\) for any injective
\(M\). Unlike classical methods designed for undirected graphs without edge labels, our
algorithms handle directed edges, edge labels, and repeated occurrences; the implementation
checks \(H=G'\) for every trajectory pair.

\begin{table}[htbp]
\centering
\caption{Summary of key symbols.}
\label{tab:symbols}
\small
\begin{tabular}{c|l}
\toprule
\textbf{Symbol} & \textbf{Meaning} \\
\midrule
$\tau$ & Complete trajectory $(q, o_0, a_1, o_1, \dots, a_m, o_m)$ \\
$q$ & Task query / instruction \\
$a_t$ & Action at step $t$ \\
$o_t$ & Observation at step $t$ \\
$r$ & Reward (success/failure) \\
\midrule
$G = (V, E, q, r)$ & Action decision graph \\
$v_i$ & Node in action decision graph (represents an action) \\
$e_i$ & Edge in action decision graph, connecting $(v_i, v_{i+1})$ \\
$\ell_n(v_i)$ & Node label (integer) \\
$\ell_e(e_i)$ & Edge label (integer) \\
$v_{\text{init}}$ & Virtual INIT node ($\ell_n=0$) \\
\midrule
$\mathcal{G}_m = (\mathcal{V}, \mathcal{E})$ & Experience Memory Graph\\
$\nu_i$ & Node in experience memory graph (one training task) \\
$\epsilon_{ij}$ & Edge in experience memory graph, connecting $\nu_i$ and $\nu_j$ \\
$I_i^{\text{intra}}$ & Intra-task insight on node $\nu_i$ \\
$I_{ij}^{\text{cross}}$ & Cross-task insight on edge $\epsilon_{ij}$ \\
\midrule
$G_i$ & Exploration graph for task $i$ \\
$G_i^*$ & Expert graph for task $i$ \\
$G_i^c$ & Common subgraph (successful workflow) \\
$G_i^e$ & Graph edit path (correction instructions) \\
$G_{ij}^c$ & Cross-task common subgraph \\
$P$ & Node matching matrix \\
$\pi$ & Soft assignment matrix (optimal transport) \\
\bottomrule
\end{tabular}
\end{table}

\begin{table*}[t]
\centering
\small
\caption{Case study on the ScienceWorld with DeepSeek-V4-Flash. Both runs share the same opening. The memory-free agent never acquires the thermometer and never performs the measurement, spending 31 of its 79 steps on rejected or no-op actions; the EMG agent executes the missing measurement chain and completes the task in 23 steps.}
\label{tab:case_study_sciworld}
\begin{tabular}{p{2.4cm}|p{6.3cm}|p{6.3cm}}
\toprule
\textbf{Phase} &
\textbf{Without Memory (reward 0.41, 79 steps)} &
\textbf{With EMG Memory (reward 1.00, 23 steps)}\\
\midrule

Opening &
\texttt{teleport to kitchen} $\rightarrow$ \texttt{look around} $\rightarrow$ \texttt{focus on thermometer} &
Identical to the left column \\
\midrule

Acquiring the tool &
\texttt{pick up thermometer} is never issued; the thermometer is left on the kitchen counter &
Step 18: \texttt{pick up thermometer} \newline {\footnotesize\color{gray}(``You move the thermometer to the inventory.'')} \\
\midrule

Reaching the substance &
29 attempts to transfer water into a container, e.g.\ \texttt{use sink on ceramic cup}, \texttt{pour sink into glass cup}, \texttt{move metal pot to sink}, repeated across kitchen, bathroom and outside
\newline {\footnotesize\color{gray}(all rejected: ``No known action matches that input'' / ``I'm not sure how to use the sink'')} &
Step 20: \texttt{focus on water}, applied to the water already present in the bathroom
\newline {\footnotesize\color{gray}(no container transfer is needed)} \\
\midrule

Measurement &
Never performed &
Step 21: \texttt{use thermometer on water}
\newline {\footnotesize\color{gray}(``the thermometer measures a temperature of 10 degrees celsius'')} \\
\midrule

Conditional action &
Not reached &
Step 23: \texttt{focus on green box}
\newline {\footnotesize\color{gray}(10\,\textdegree C is below the 150\,\textdegree C threshold set by the task)} \\
\midrule

Outcome &
31 of 79 steps rejected or no-ops; step budget exhausted &
7 of 23 steps ineffective; task completed \\

\bottomrule
\end{tabular}
\end{table*}

\section{Discussion on the Robustness of Graph Edit Path}

We further analyze the robustness of our graph edit path extraction (Algorithm~\ref{alg:derive_edit}). Our graph matching procedure relies on the FGWAlign solver~\cite{26}, which may not always identify the globally optimal node correspondence. According to the original FGWAlign evaluation, the recall of the predicted edit path against the ground-truth shortest path is around 90\% for graph pairs with tens of nodes. Importantly, a recall below 100\% does not necessarily indicate an invalid edit path; it may instead reflect redundant edit operations.

Our algorithm provides a guarantee that is independent of the optimality of the node matching: for any injective matching (M), applying the extracted edit operations to the source graph yields the target graph (G’). Thus, even when the matching is suboptimal, the resulting edit path remains structurally valid, although it may contain more operations than the shortest possible path. This distinction allows us to separate the correctness of graph transformation from the optimality of the matching. The empirical recall reported for FGWAlign further suggests that the solver often preserves a substantial portion of the optimal edit path.

This property also provides a fallback when graph matching is imperfect. In the extreme case of an entirely incorrect matching, the edit path may effectively replace the failed trajectory with the expert trajectory, offering a valid alternative. So our algorithm guarantees a valid graph transformation and retains a fallback toward the successful trajectory.

\section{Notations}
Details are available at Table~\ref{tab:symbols}.

\section{Case Study: EMG helps avoid potential error}

Table~\ref{tab:case_study_sciworld} compares DeepSeek-V4 Flash on the ScienceWorld with and without the
memory retrieved by EMG. Both runs begin identically and diverge on one point: whether the
agent ever acquires the thermometer and applies it to the target substance.

The memory-free agent treats \texttt{focus on thermometer} as having secured the instrument and
then commits to an unsupported sub-plan, namely transferring water into a container before
measuring it. It issues 29 such attempts --- \texttt{use sink on ceramic cup},
\texttt{pour sink into glass cup}, \texttt{move metal pot to sink} and variants, across the
kitchen, the bathroom and the outside location --- none of which exist in the action space; the
environment answers ``No known action matches that input'' or ``I'm not sure how to use the
sink''. In total 31 of its 79 steps are rejected or produce no state change. It never issues
\texttt{pick up thermometer}, \texttt{focus on water} or \texttt{use thermometer on water}, and
the run ends with the step budget exhausted at reward 0.41, the partial credit for the
\texttt{focus} checkpoints alone.

The node insight retrieved for this task was derived by graph matching on a \emph{different}
substance, a training pair for measuring the melting point of lead, and states verbatim:

\begin{quote}
\emph{The edit path correctly highlights that the failed trajectory is missing all heating and
measurement steps (e.g., \texttt{activate blast furnace}, \texttt{use thermometer on liquid
lead}). \dots Corrected reasoning: the instruction ``focus on the lead'' is a subgoal, not the
final objective. The agent must proceed to: 1) acquire the thermometer and lead, \dots}
\end{quote}

This is precisely the error that the memory-free agent commits on the water task. Guided by it,
the EMG agent picks up the thermometer (step 18), focuses on the water already present in the
bathroom (step 20), measures it (step 21), and selects the box implied by the reading
(step 23), reaching reward 1.00 in 23 steps. 

\end{document}